# High-temporal-resolution event-based vehicle detection and tracking

Zaid El Shair,[a,*] Samir Rawashdeh[a]

[a]University of Michigan-Dearborn, Department of Electrical and Computer Engineering, 4901 Evergreen Rd, Dearborn, Michigan, USA, 48128

**Abstract**. Event-based vision has been rapidly growing in recent years justified by the unique characteristics it presents such as its high temporal resolutions (~1us), high dynamic range (>120dB), and output latency of only a few microseconds. This work further explores a hybrid, multi-modal, approach for object detection and tracking that leverages state-of-the-art frame-based detectors complemented by hand-crafted event-based methods to improve the overall tracking performance with minimal computational overhead. The methods presented include event-based bounding box (BB) refinement that improves the precision of the resulting BBs, as well as a continuous event-based object detection method, to recover missed detections and generate inter-frame detections that enable a high-temporal-resolution tracking output. The advantages of these methods are quantitatively verified by an ablation study using the higher order tracking accuracy (HOTA) metric. Results show significant performance gains resembled by an improvement in the HOTA from 56.6%, using only frames, to 64.1% and 64.9%, for the event and edge-based mask configurations combined with the two methods proposed, at the baseline framerate of 24Hz. Likewise, incorporating these methods with the same configurations has improved HOTA from 52.5% to 63.1%, and from 51.3% to 60.2% at the high-temporal-resolution tracking rate of 384Hz. Finally, a validation experiment is conducted to analyze the real-world single-object tracking performance using high-speed LiDAR. Empirical evidence shows that our approaches provide significant advantages compared to using frame-based object detectors at the baseline framerate of 24Hz and higher tracking rates of up to 500Hz.

*Zaid El Shair, E-mail: zelshair@umich.edu

## 1 Introduction

In the last couple of years, neuromorphic event-based vision has been gaining attention in the literature and growing exponentially[1,2]. Event-based sensors, first introduced in 2008[3], propose a novel type of sensing modality with distinct and advantageous characteristics compared to the typical frame-based cameras. These sensors, commonly referred to as *event cameras*, capture brightness changes asynchronously and independently per each pixel of the sensor's pixel array. Each of these captured brightness changes is known as an *event*. Every event consists of 4 different types of information including a microsecond-resolution timestamp $t$ of when it was detected, an $x$ and $y$ pixel coordinates at which the event has occurred, and a polarity $p$ indicating the type of brightness change that was registered (i.e., positive or negative). Accordingly, an event is defined as $e = \{t, x, y, p\}$.

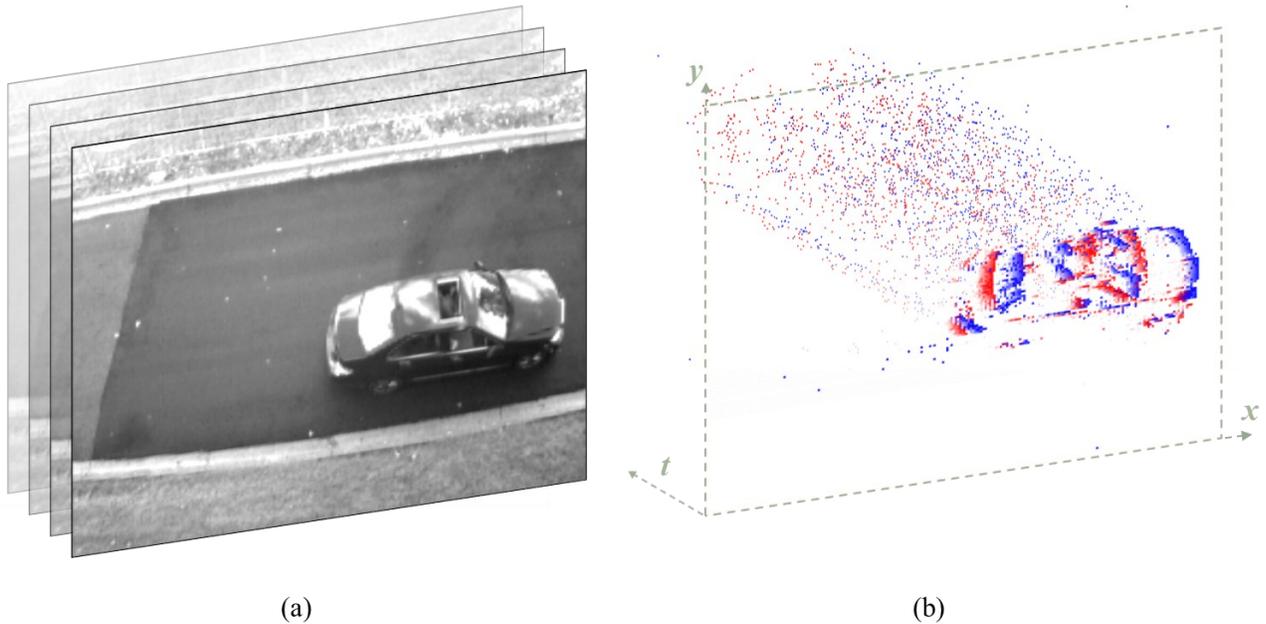

(a)                                                 (b)

**Fig. 1** Comparison between the frame-based and event-based modality output, resembled by (a) the synchronous images captured by a frame-based camera, and (b) the asynchronous events captured by an event-based sensor. Notice how the events are mainly generated around the edges of the moving vehicle, where brightness changes exist), in contrast with the images captured by the frame-based camera where the static background is redundantly captured across consecutive frames at a fixed sampling rate.

In contrast, conventional frame-based cameras capture images synchronously at a fixed rate (typically ~30 frames per second), recording the color intensity of each pixel, regardless of whether there were any changes, in every frame generated at a fixed sampling rate. This causes frame-based cameras to be very susceptible to producing redundant data that may resemble a static background in a given scene, especially when the camera is stationary (e.g. undergoing limited, to no, ego-motion). Meanwhile, event cameras would mostly capture changes in the scene, often resembling motion, at the instances of their occurrence. Nevertheless, event cameras can be less effective in scenes of limited to minimal motion, where there would be a lack of visual signal to reliably utilize this modality on its. Thus, causing the unimodal event-based implementations to possibly be unreliable in some scenarios. Overall, we illustrate the difference between the visual data output of the two modalities (i.e., frame-based and event-based) in Fig. 1.

The main specifications of a typical event camera include a very high temporal resolution (~1 µs per event making it robust to motion blur), low latency in the order of microseconds, and a high dynamic range (HDR) of over 120dB (compared to ~60dB of conventional frame-based cameras) while requiring considerably less power[2,4]. Given these properties, event-based vision proposes an exciting domain with great promise if explored and applied properly. Current works have utilized these specifications for different applications such as motion deblurring[5], high-framerate HDR video synthesis[6], image reconstruction from events[7], and enhanced object detection[8] to name a few.

While the potential behind this novel visual-sensing technology is evident, we believe that it can provide optimal benefits when incorporated with frame-based vision, as both modalities can be complementary to each other when they are correctly utilized. Such an approach can enable a more robust perception performance for different automated applications.

In this work, we explore a combined frame- and event-based approach for vital computer vision tasks, namely object detection and tracking. Object detection is an essential component for automated systems to provide awareness of the surroundings at a given instant. Meanwhile, object tracking enables the system to associate these detections across time, supplying the temporal element for its interpretation of its surroundings. Both components, although challenging and processing-intensive, are critical for a complete and reliable system perception performance, as they assist in different tasks, such as motion planning and obstacle avoidance, and play an important role in various applications in robotics, including traffic monitoring and surveillance systems[9], and autonomous vehicles[10,11].

Object detection performance can vary based on the method used. Deep Neural Network (DNN) based object detectors have recently dominated the state-of-the-art[12–16], thanks to the unparalleled advancements in deep learning in general[17], and the emergence of deep convolutional neural networks (CNN) specifically[18,19]. Nevertheless, the different implementations in the literature are often constrained by a trade-off between object detection accuracy and latency. YOLOv3[12], for instance, offers real-time inference speeds, however, at the cost of lower accuracy and possibly more inconsistent performance, which could affect the overall object tracking performance due to the intermittent detections. On the other hand, FasterRCNN[16] offers better object detection accuracy which the object tracking framework can benefit from, however, at the expense of considerably higher latency, making it less ideal for real-time detection and tracking systems.

Similarly, the temporal resolutions of frame-based object detection and tracking can be limited by the framerate of the input source which is typically fixed, such as a camera that typically has a low output framerate. This sets an upper bound for the resolutions of object tracking, given that interpolation techniques are excluded which are not beneficial for a real-time online system. Furthermore, even if a higher framerate source is used along with a DNN-based object detector, the system's operational latencies would be further impacted, imposing stringent hardware requirements to be able to achieve real-time computation performance.

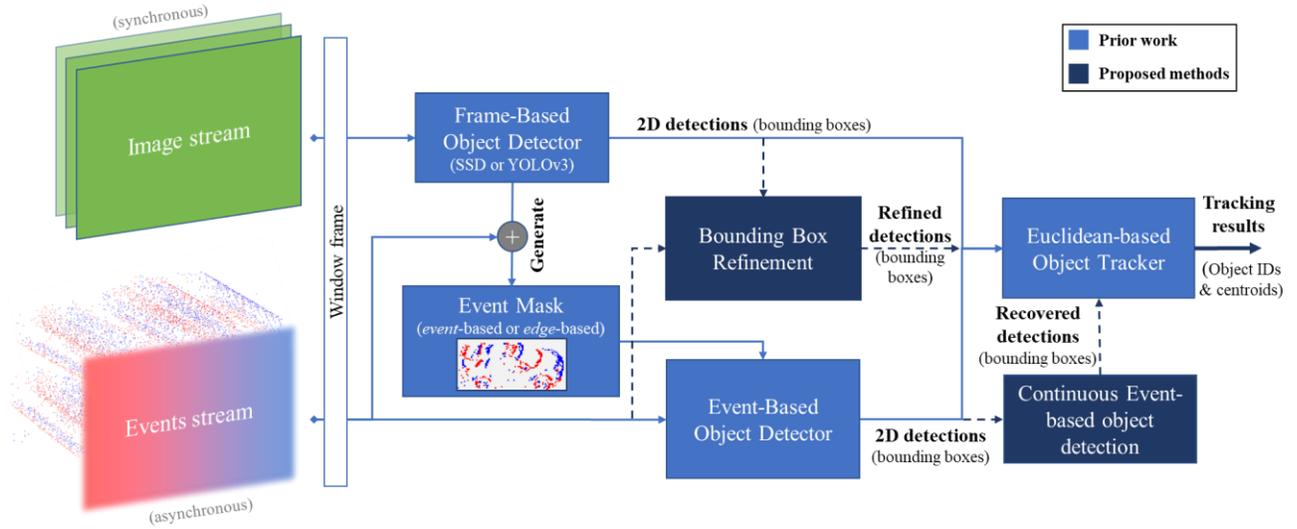

**Fig. 2** Block diagram demonstrating the hybrid multi-modal object detection and tracking framework with the proposed methods, which include bounding box refinement and continuous object detection using event-based methods. Visual streams from both modalities are synchronized and passed through the framework to yield multi-object tracking results.

In this paper, we extend and improve on our prior work[20], which explored the feasibility of high-temporal-resolution object detection and tracking using a hybrid multi-modal approach that incorporates synchronized image and event data, by presenting two additional methods that improve the overall object detection and tracking performance using event-based techniques. First, we improve the precision of bounding boxes (BBs) proposed by frame-based object detectors using a combination of event data and classical computer vision methods. Second, we enhance the robustness and consistency of frame-based object detectors using event-based detection methods. This method is automatically initiated whenever the frame-based object detector fails to detect an object in a given frame, thus improving the object detection reliability and the corresponding tracking performance by leveraging the high-temporal-resolution event data. Third, we numerically assess the effects of these methods under different frame-based object detection models using a fully labeled dataset (at multiple tracking rates) along with state-of-the-art multi-object tracking (MOT) metrics, including the higher order tracking accuracy (HOTA) metric for evaluating multi-object tracking. This is followed by a simple computational cost analysis for the presented event-based methods in comparison with the frame-based components. Finally, we validate our work with a real-world experiment using a high-speed LiDAR to provide high-temporal-resolution positional measurements of a vehicle being tracked. The general framework of this work along with the proposed methods are demonstrated in Fig. 2.

Overall, the main contributions of this paper are summarized as follows:

- We present an event-based bounding box refinement method for static scenes, and an event-based method for recovering undetected objects in the frame domain, which improves the object detection and tracking performance compared to the frame-based baseline.
- We present an ablation study that quantitively verifies the benefits of each introduced event-based method and their optimal combination using the HOTA metric.
- We provide a computational latency analysis for the introduced methods as well as the core components of the proposed system.
- We perform a real-world validation experiment using a high-speed LiDAR that evaluates how well the presented framework, including the additional event-based methods, estimates the vehicle position at different temporal resolutions and tracking rates.

## 2 Related Work

*2.1 Frame-based Approaches*

Frame-based object detection and tracking methods have had significant developments throughout the last decade. The advancements of DNN-based object detectors have enabled a very robust object detection performance[12–16], which is complemented by various data association techniques to achieve multi-object tracking (MOT)[21,22]. This approach of object tracking is commonly referred to as *tracking-by-detection*.

When it comes to object detection, deep learning-based object detectors are categorized as either one-stage[12–15] or two-stage detectors[16]. While two-stage detectors are often more accurate and robust, one-stage detectors sacrifice some accuracy for inference speeds making them more appropriate for real-time systems but with degraded performance. One-stage object detectors include YOLOv3[12], SSD[13], RRC[14], and RetinaNet[15], whereas two-stage object detectors mainly include FasterRCNN[16].

As for object tracking, State-of-the-art MOT implementations are designed to operate in either an *online* or a *global* manner. Online methods are more appropriate for real-time robotic systems compared to global methods[23,24] which require full knowledge of all the current and future data (more suitable for other types of applications). The output of these implementations is typically evaluated using well-known object tracking metrics such as the CLEAR MOT[25] or the more recent HOTA[26] metrics on various MOT benchmarks such as MOT20[27].

Analogous to the literature, we leverage a tracking-by-detection approach to enable multi-modal object detection and tracking using image frames as well as events. Focusing on the object detection aspect, our combined approach accomplishes this by the use of pre-trained, one-stage, frame-based object detectors, specifically YOLOv3[12] and SSD[13], while employing a simple data association metric, which is Euclidean distance[28]. The choice of this data association metric is based on the assumption that the resulting detections, ideally, should be continuous given the high-temporal-resolution nature of event data. Further, using methods discussed later in this paper, frame-based object detections are used as the basis for enabling the detection and tracking of the objects in the scene in between frames at varying temporal resolutions, using the event data.

*2.2 Event-based Approaches*

On the other hand, event-based object detection and tracking methods, compared to the frame-based implementations, are in their preliminary stages and have yet to utilize the event data to its full potential. Unlike the frame-based domain, event-based vision has had varying approaches when it comes to either object detection or object tracking. For instance, single-modal event-based object detection is currently in the experimental phase, where learned methods make up the majority of the state-of-the-art. Learned implementations typically combine and embed events in an image-link representation which are used with modified frame-based DNN architectures to make them compatible with event data, in either a temporal[29–31] or a non-temporal[32] manner. Recurrent and temporal approaches are typically more suitable for event data, given that events only provide brightness changes and not absolute brightness, at a given point in time, in contrast to frames. Thus, the temporal approaches would incorporate meaningful input history, instead of treating each split, of the input stream, independently, yet at the expense of higher computation. Although the single-modal event-based approaches are promising, their performance typically lags behind both the frame-based solutions, under normal conditions, as well as the combined solutions which incorporate both modalities[8,29,33–35]. The main reason behind this is due to the nature of the event data, especially in scenes where there is limited motion. Besides, the lack of enough labeled datasets, available for training deep learning models, only increases the performance gap between both domains. In our work, we present a multi-modal approach that leverages pre-trained frame-based object detectors to initiate objects in the scene. Afterward, these detections are used to generate templates for each object, which we refer to as event masks, in order to detect objects at varying tracking rates using the asynchronous and high temporal resolution event data temporally. Moreover, unlike fully learned approaches, a mixed approach of learned and classical methods can be more computationally efficient when requiring very high tracking rates. Furthermore, a classical and designed detection method does not require large datasets of labeled data for implementation, which is still a constraint in the event-based vision domain.

As for event-based object tracking, most works focus mainly on the detection aspect with typically a single-object tracking approach. One common approach is the use of clustering techniques[36–39]. Clustering events is a low-cost and intuitive way for simple object tracking applications enabled by the nature of events that often resemble movement, as they would mainly be generated around the objects that are in motion. Nevertheless, clustering can be prone to object collisions and does not apply to object classification. Conversely, non-clustering methods have explored various single and multi-modal object detection and tracking techniques[40–44]. For example, Chen et al.[43] proposed an event-to-frame conversion algorithm to enable an asynchronous tracking-by-detection approach. Meanwhile, Ramesh et al.[44] presented an online object tracking framework with a moving event camera using a local sliding window technique, with a global object re-identification using an event-based object detector whenever the tracker loses the object.

Overall, we notice that few event-based object tracking works focus on real-world objects such as vehicles, where most use data of shapes in indoor scenes, and approach this problem from a single object tracking aspect (i.e., without the use of well-defined MOT metrics like in the frame-based domain). A primary constraint behind this is the limited number of labeled object detection and tracking event-based datasets available publicly. Our prior work[20] provided the first fully labeled event-based dataset conforming to the typical MOT standards[45], including object IDs, unique to each object's trajectory, along with two-dimensional (2D) BBs provided at multiple temporal resolutions and tracking rates.

To summarize, our previous work[20] focused on exploiting the high-temporal resolution of the event data to enable high-temporal-resolution object detection and tracking using a combined approach that utilizes learned frame-based object detectors and classical event-based methods. Our proposed framework was evaluated using a labeled vehicle dataset with state-of-the-art MOT metrics[26]. However, our work was constrained by the performance of the frame-based detectors to initiate detections and redetect objects at every new image frame. In this work, we propose two methods that help boost the detection performance, where we use event-based techniques to improve the accuracy of the generated BBs and to support object detection when a frame-based object detector fails to detect an object, previously tracked, in a given image.

## 3 Methodology

In this section, we initially summarize the hybrid framework presented in our prior work[20] in more detail, then describe some of its limitations and the motivation behind the additional methods introduced in this paper (Sec. 3.1). Afterward, we present and break down the proposed event-based methods to improve object detection and tracking (Sec. 3.2 and Sec. 3.3). Finally, we perform an ablation study to assess the effectiveness of these methods and figure out the optimal combination to use (Sec. 3.4).

### 3.1 High-Temporal-Resolution Object Detection and Tracking Framework

In our prior work[20], we have introduced a hybrid approach, that leverages both frame-based and event-based vision modalities to enable high-temporal-resolution object detection and tracking (as demonstrated in Fig. 2). We break down the framework in the following subsections.

#### 3.1.1 Parsing the multi-modal streams using window frames

To incorporate the data streams of both modalities, the temporally synchronized image and event streams are divided into a moving *window frame* containing any available image frames as well as a predefined interval of event data (e.g., 50 ms). Based on the desired tracking rate $k$ Hz, the window frame will move forward $\frac{1}{k}$ ms for every step, with an interval size $\Delta t$. For example, given a tracking rate $k$ equal to 200 Hz and $\Delta t$ set as 50 ms, the window frame will take steps of 5 ms while containing the latest 50 ms of event data. Note that inter-frame event-based object detection is executed only if the desired tracking rate $k$ is set higher than the framerate of the images (ideally $\leq \frac{k}{2}$ Hz).

*3.1.2 General multi-modal detection and tracking framework*

Following a tracking-by-detection approach, pre-trained frame-based object detectors (i.e., YOLOv3 and SSD) are used to detect objects in the image domain (when available in the frame) and initiate their tracking. Meanwhile, an event-based, inter-frame object detection method is used to detect previously detected objects in the event domain, in the blind time between consecutive frames. Accordingly, whenever a window frame containing an image is read, the frame-based detections are used to initiate the object tracker with these objects by associating each with a unique object ID. At the same time, the resulting two-dimensional BBs of the detected objects, representing their location relative to the frame, are used to generate templates that are later required for the event-based object detection process. We refer to these templates as *event masks*. Note that the event masks are of the same size as the objects detected in the image frames. Afterward, these event masks are used to enable inter-frame object detection in the subsequent window frames that only contain events, until a window frame containing an image is read at which point this process is repeated. Throughout this process, the detection results of each window frame are fed into the object tracker, which associates the latest detections by initiating any new objects that entered the scene, updating the position of previously tracked ones, and removing the objects that have left the scene. Moreover, we utilize a simple Euclidean distance[28] minimization function to associate recent detections with the objects currently being tracked. This simple association metric was chosen to give more emphasis on the cooperative multi-modal object detection process and is motivated by the expected continuous tracking results due to the high temporal resolution of the event data.

*3.1.3 Event-based object detection using event masks*

Here, we break down the event-based object detection process referred to in our framework. Given a set of currently tracked objects with corresponding event masks, the event-based object detection is achieved using a sliding window mechanism of matrix multiplication between each object's event mask and the events that are available in the current window frame and located within the confined search region. The search region is generated around the latest detected position of each object, albeit with a larger area to cover all possible displaced positions. The results of all the possible combinations of this sliding window mechanism are stored in a cost matrix. Afterward, the highest value in the cost matrix, which resembles the best correlating position, is used as the object's inter-frame position, as long as that value meets or exceeds the preset score threshold. Otherwise, no detection is generated for the corresponding object within that window frame. Therefore, resulting in a missed inter-frame detection which creates gaps in the estimated trajectories of the tracked objects.

Note that the event-based object detection process is only applied to the objects that were previously detected in the most recent window frame that contained an image using the selected frame-based object detector. Thus, if an object was not detected in the latest image in the data stream, the event-based object detection process is not initiated until the object is detected in a future image that precedes a window frame of events. Further, we note that the events in each window frame are temporally weighted, giving the highest weight to the newest events and lower weights for older events. By giving more significance to the newest events, which typically resemble the latest movements of the object, this process was found to improve event-based object detection precision[20] and is the applied practice throughout this paper.

We have also introduced and evaluated two different types of event masks (i.e., event-based and edge-based) which are demonstrated in Fig. 3. Event-based masks (shown in Fig. 3(a)) are generated by accumulating all of the events located within the 2D BB of the detected objects using the frame-based detectors. The event-based masks represent the object with the most recent event in each pixel while retaining their polarity (as +1 for positive events and as -1 for negative events). Thus, resulting in a sparse 2D matrix of integers containing integer values of 0, +1, and -1. Event-based masks are ideal when there is limited motion in the background of the object, causing the events generated to be mostly due to the object itself. However, it is vulnerable to events generated due to noise (global shutter or shadows) and is not ideal when the detected object (in the image) is not in motion, thus limited events are generated that can be used to generate the mask. On the other hand, edge-based masks, shown in Fig. 3(b), rely on the actual image crop representing the object's BB. This crop is processed and converted to a binary mask representing the edges of the object, albeit without the polarity information where absolute values are used instead. Thus, the edge-based event mask is represented by a sparse 2D matrix of integers containing values of 0s and +1s. The edge-based mask mimics the events generated by a moving object which are typically around its edges. Therefore, possibly making it a more appropriate choice for an object that is rapidly changing its direction. However, it is susceptible to any edges in the background and any distortions in the image due to motion blur and poor dynamic range. Overall, despite their limitations, both types of event masks can be improved by different methods, such as filtering, or generating learned masks instead of handcrafted ones. However, that is not the focus of this research but can be addressed in future works. Nevertheless, they are evaluated further in this work, in addition to the proposed methods described in the following sections.

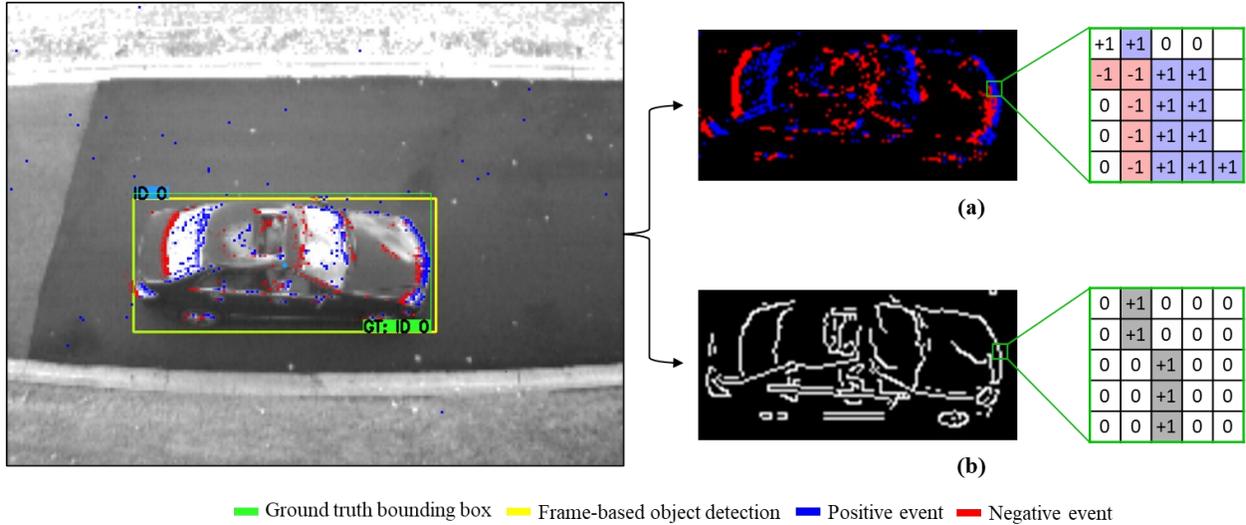

**Fig. 3** Demonstration of the two different event mask types and how they are generated after the object is initially detected in the frame, including (a) event-based mask, and (b) edge-based mask. The frame-based object detector used in this example is SSD.

*3.1.4 Limitations of the framework*

In summary, our prior results have proved the ability to generate high-temporal-resolution object tracking and detection by utilizing event data in combination with images. By incorporating the temporal information of the events while using a sufficient history of event data (50ms), higher temporal tracking rates (up to 384Hz demonstrated) were feasible with very minimal performance deterioration compared to when tracking at the base framerate (24Hz) using only image frames. By using classical computer vision techniques to amplify the tracking rates, minimal overhead is added, in comparison with the use of learned methods (such as frame-based object detectors) at higher rates.

While our general framework has shown robust results of higher temporal resolution tracking from a low sampling-rate input source with the aid of event data, it is limited by the relatively inconsistent performance of frame-based object detectors that are optimized for real-time performance (such as YOLOv3[12]). This is exemplified in both the poor BB alignment accuracy and the commonly missed detections of objects in the scene (i.e, false negatives). Poor BB estimation (also known as localization accuracy) reduces the overall detection and tracking performance, which is evident by the tracking metrics used. Likewise, missed detections cause fragments in the estimated trajectories of the tracked objects, especially as our presented framework is reliant on detecting the object in the frame domain in order to initiate the object detection process in the event domain in subsequent window frames of event data. Therefore, in this work, we present two additional methods that should improve the performance consistency of object detection (in any given frame), and the precision of the generated bounding box that highlights the position of the detected objects. Our prior approach along with the proposed methods are highlighted in Fig. 2. Finally, we quantitatively verify the advantages of these enhancements with the use of state-of-the-art multi-object detection and tracking metric, HOTA[26], presented in the form of an ablation study in Sec. 3.4.

### 3.2 Event-based Bounding Box Refinement

Here, we explore an intuitive event-based method for BB refinement, inspired by the frame-based BB refinement methods that have been previously proposed in the literature[46]. This method is broken down into two main stages, which are event extraction and three-dimensional (3D) matrix generation, as well as BB filtering and refinement.

#### 3.2.1 Event extraction and 3D matrix generation

Once a new window frame containing an image is loaded, the frame-based object detector generates predictions, each including object classes and 2D BBs. Similar to the event mask generation process (described in Sec. 3.1), the events available in the mask's area within the search region available in the current window frame, are extracted and added to a two-dimensional matrix. Note that the BB of the mask is a slightly enlarged version of the initial object detection's BB, thus

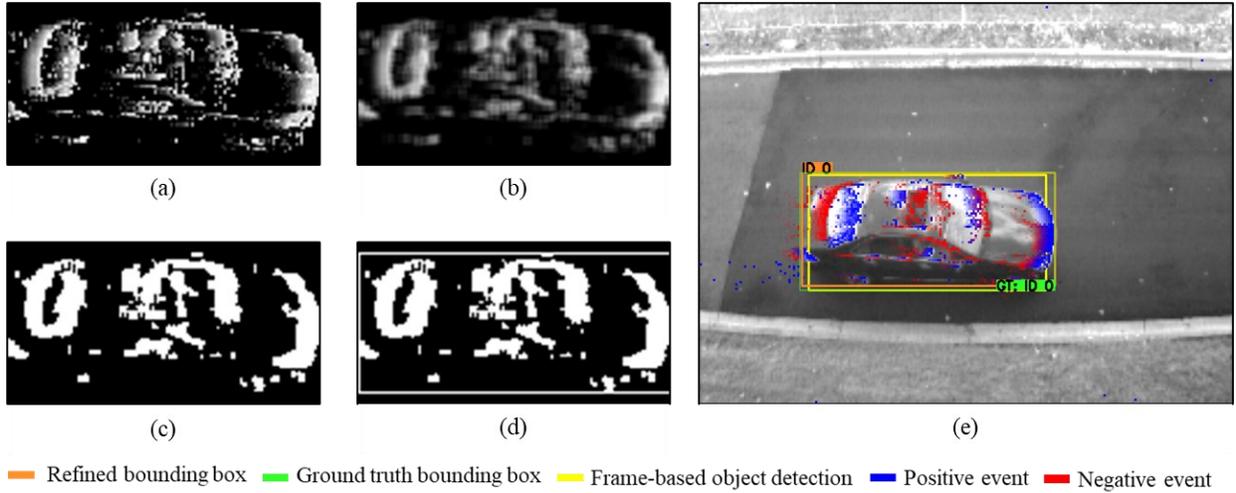

▬ Refined bounding box ▬ Ground truth bounding box ▬ Frame-based object detection ▬ Positive event ▬ Negative event enabling a larger initial area to freely enlarge or reduce the BB as needed based on the true size of the object. This is because frame-based object detectors might either over- or under-estimate the size of the detection's BB. Afterward, a depth channel is added to convert the resulting matrix into three dimensions (from 2D) to be able to process it like a typical image. This is followed by removing the polarity information by finding the absolute values of the events present in the matrix (negative events and positive events are treated alike).

**Fig. 4** Overview of the bounding box refinement process of the object detected using event data. (a) Events are combined to form a grayscale image, (b) then the image is filtered to remove noise using an average blur. Finally, the resulting blurred grayscale image is then thresholded to generate a binary version, (d) where a best-fit bounding rectangle is formed that highlights the object more precisely. (e) Results are demonstrated, where the frame-based object detection and the refined BB are highlighted by the yellow and bright orange BB, respectively, while the ground truth label is shown in green.

Given that our approach incorporates temporal information, we linearly weight each event (in the 3D matrix), giving more weight (higher value) to the more recent events at that tracking time instant, similar to our approach of weighing the window frame's events temporally discussed earlier. Finally, the values of each entry in the 3D matrix are normalized to values between 0 and 1, then converted to a range of 0 and 255 representing a grayscale image as shown in Fig. 4(a).

*3.2.2 Bounding box filtering and refinement*

Subsequently, the second stage is then applied, which involves filtering the resulting grayscale image and generating a refined BB.

The resulting image from the previous stage is initially filtered by blurring it using an averaging filter with a kernel size of 3x3, as shown in Fig. 4(b). Blurring an image is a standard method used to smoothen an image and remove noise. In our use case, we adapt this method to the grayscale image representing the events to assist in removing the ones generated due to noise, such as shadows or global shutter which is common in hybrid cameras.

The resulting blurred image is then thresholded and converted to binary values (0 or 1 per each matrix's values) using Otsu's method[47] to differentiate the object itself from the background or any noise. Otsu's method was noticed to help find a more optimal global threshold value to binarize an image appropriately which is shown in Fig. 4(c). Finally, a best-fit BB is then generated that would precisely cover the whole object by providing the minimum contour possible to fit the object, as shown in Fig. 4(d). The resulting BB is then fed into the object tracker to associate the detected object with ones previously tracked based on Euclidean distance[28]. Fig. 4(e) shows the detected frame-based BB (produced by YOLOv3[12]) against the refined and the ground truth BBs at a given time instant. We can observe that event-guided refined BB matches the ground truth more accurately than the initial frame-based detection. Note that in our prior work, the BBs generated by the frame-based object detectors are used as is without any refinement or modification, while the subsequent, inter-frame detections, would retain the same BB size as well, based on the framework described in Sec. 3.1. Finally, this method is also applied in subsequent window frames of event data to enable BB refinement at higher temporal resolutions as well when detecting objects using the generated event masks. Thus, the BB of the detection is continuously and dynamically refined at each tracking instant.

Undoubtedly, we note that an insufficient number of events can adversely impact the outcome of this process, possibly leading to a very small, and inaccurate, BB. Therefore, it is important to set a simple check to initiate this process. Accordingly, we set a minimum threshold for the sum of the temporally weighted events, within the initial BB, to permit the refinement process. This was applied throughout our work and is reflected in the results presented in Sec. 3.4.

Moreover, we have observed that this method can be negatively affected by more recent events generated due to noise, which are the result of some shadows or shutter noise as discussed earlier. Newer events, relative to the tracking time instant, are given more weight, and thus are not always filtered out by our method. This leads these events to affect the accuracy of the best-fit BB generated in the last step. Similarly, we have noticed that more precise models, such as SSD[13], can be adversely impacted by this process, mainly due to the same type of noise events, besides the fact that the generated detections have very precise BBs, to begin with, as observed in our work.

*3.3   Continuous Event-based Object Detection and Recovery*

Efficient object detector models, such as YOLOv3, also suffer from an intermittent object detection performance. An object that appears in many consecutive frames might be detected in some but missed in others. This can affect the overall performance of the object detection and tracking given our framework's dependency on the frame-based object detector to be able to initial the inter-frame detection, as evident in our prior work's results[20]. Meanwhile, more sophisticated models, such as SSD[13], have more consistent performance, albeit at the expense of more computational requirements and greater latencies. Accordingly, to minimize the effects of intermittent frame-based object detection performance, we present a method that recovers missed detections and false negatives of objects that were previously tracked, using event data.

As described in Sec. 3.1, an event mask is generated whenever an object is detected using the frame-based object detectors. Each event mask is then used in a sliding-window mechanism to find the object's optimal inter-frame position within an enlarged area known as the search region. The highest correlating position is then assumed to be the inter-frame object position, hence the resulting event-based object detection. Nonetheless, in our prior work[20], this method was only integrated whenever an object is successfully detected in a given image to be able to perform event-based object detection before the subsequent image is read. If an object was missed, then it is no longer tracked till it is detected again by the frame-based object detector with possibly an incorrect object ID (based on the association technique). Therefore, in this work, we evaluate a similar application of using the event masks, but of previously tracked objects, to recover any missed detections at a given image frame. Thus, limiting the gaps in the estimated objects' trajectories and improving the overall object detection and tracking performance.

We describe this process as follows. Given a set of previously tracked objects, an undetected object in the image, within the latest window frame, is marked as "disappeared" after associating all of the frame's detections using the object tracker. The missed object's latest event mask, utilized in the previous window frame, is similarly used to initiate the event-based object detection process using the events available in the latest window frame, located within the search region. This follows the same process utilized in our prior work for inter-frame event-based object detection, as described in Sec. 3.1. Nevertheless, due to lower certainty, a significantly higher event-based detection score threshold is used. This threshold resembles the minimum matrix multiplication summation value that must be met or surpassed in order to consider the object detected in the event domain. We apply this validation process to limit the detections that may result from events produced by other objects or generated by noise, given the lower detection confidence to begin with, and considering that this method is neither dynamic nor learned, unlike the frame-based object detectors we use in our framework.

As with the former method, this one has its limitations and drawbacks as well, if not properly optimized. This method can be negatively affected by several factors, such as false positives (incorrect detections) or poorly aligned detections generated by frame-based object detectors, as shown in Fig. 5(a), which are given higher precedence and are prioritized over the event-based methods, given their presumed robust and dynamic performance. These false positives, even if intermittent or sporadic, would be used for continuous event-based detection, as described earlier, to recover the missed detections. Based on the event mask generation process, these false detections can lead to consistent and continuous false detections in subsequent window frames if subjected to a sufficient number of events generated due to noise or other external factors, as shown in Fig. 5. Thus, affecting the overall detection and tracking accuracy.

*3.4 Ablation Study*

To verify the feasibility and the benefits of the proposed methods, we conduct an ablation study to assess the influence of each on the overall object detection and tracking operation and to indicate the right combination of methods to achieve optimal performance. Moreover, we provide a computational latency analysis of the estimated average latency of each core component of our framework.

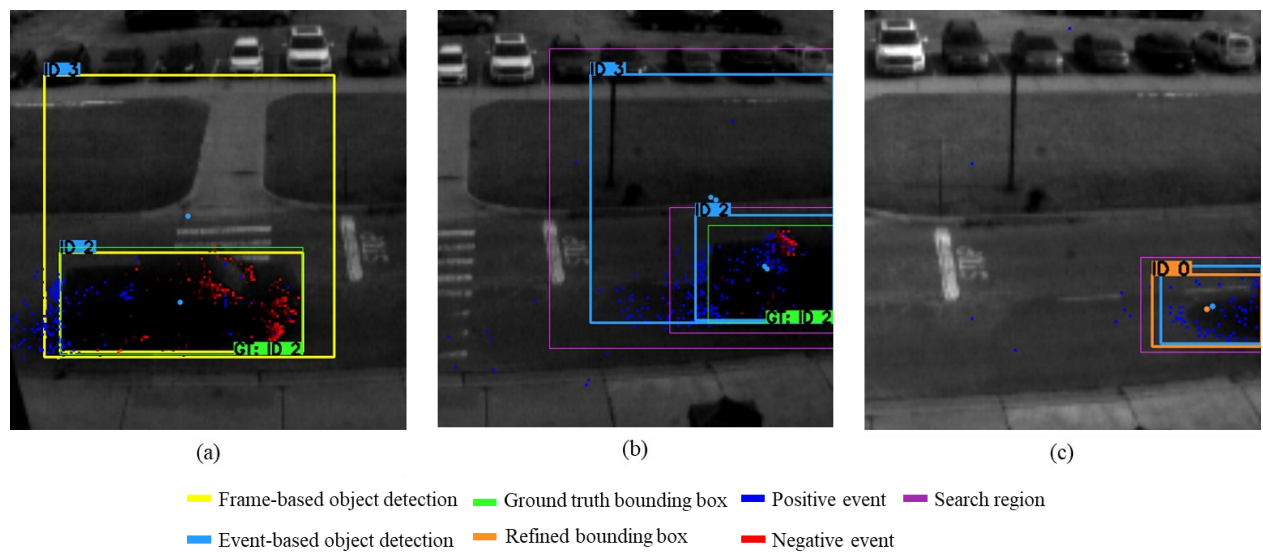

**Fig. 5** Demonstration of some failure modes when accompanied by continuous event-based object detection and bounding box refinement: (a) false object detection generated by the frame-based object detector SSD, (b) which are then continuously yet incorrectly detected using events in subsequent window frames; (c) an object is falsely detected and recovered due to events generated by the shadow of a vehicle leaving the scene.

*3.4.1 Evaluation parameters and configurations used*

Similar to the setup used in our previous work[20], we use two pre-trained frame-based object detectors which are YOLOv3[12] and SSD[13]. We select the YOLOv3-320 variant trained on Microsoft COCO labels[48], with an input image size of $320 \times 320 \times 3$; and the VGG19-based SSD-300 variant, trained on PASCAL VOC labels[49], with an input image size of $300 \times 300 \times 3$. The models selected propose a trade-off between latency and accuracy. Moreover, we set both confidence and non-maximal suppression thresholds are set to 50%. Only predictions related to the 'vehicle' class (and its different forms) are considered, whereas other object classes are simply ignored and filtered out. Both frame-based object detector models are used along with different detection and tracking configurations described in this section.

Furthermore, we build on the two best approaches presented earlier[20] with the proposed methods discussed in Sec. 3.2 and 3.3, to verify the validity of our assumptions. The two best approaches incorporated a moving window frame of image and event data, of the last 50 ms which are temporally weighted, at any given tracking instant, while only varying by the use of different event mask types (event-based and edge-based described in Sec. 3.1.3) for event-based object detection. These modes are referred to as modes 2 and 3, for event-based and edge-based masks, respectively. The combinations relating to each of these modes are referred to as A(1-3) for mode 2, and B(1-3) for mode 3. Moreover, we provide the results of the single-modal, frame-based approach for object detection and tracking (without any event-based methods), to provide a baseline reference to compare with the proposed improvements at the preset rate of 24Hz.

As for the dataset used, we evaluate the different tracking configurations using the same labeled dataset presented in our prior work[20]. A total of 63 sequences were captured by the hybrid sensor DAVIS 240c[4], with fully labeled vehicle objects (BB and object ID for each) at different tracking rates (24, 48, 96, 192, and 384Hz) resembling various temporal resolutions. DAVIS 240c consists of both an active pixel sensor (APS), and a dynamic vision sensor (DVS), which are frame-based and event-based sensors, respectively, that use the same pixel array. The APS captures grayscale images at 24 frames per second (or at a sampling rate of 24Hz), whereas the DVS captures events asynchronously at a resolution of 1μs. The data was captured while the camera was static (i.e., no ego-motion applied) pointing downwards at the street with multiple vehicles driving by at varying velocities and acceleration rates.

Finally, the well-defined object detection and tracking metric HOTA[26] is used to evaluate the overall object detection and tracking performance of the different tracking configurations. Compared to prior metrics in the literature, the HOTA metric provides a good balance between the overall object detection, association, and localization accuracy in a combined metric. In addition, HOTA can be decomposed into a series of sub-metrics that describe each. These sub-metrics include the Detection Accuracy (DetA), which describes how well detections are aligned; the Association Accuracy (AccA) which measures how well matched-object trajectories are aligned and associated across time; and the Localization Accuracy (LocA), which refers to how well spatial alignment is between the predicted and ground truth detection. The main metric HOTA, as well as the sub-metrics, are calculated over a range of intersection-over-union threshold α values, spanning from 0.05 to 0.95 with increments of 0.05. The HOTA metrics are described in further detail in

**Table 1** Results of the ablation study on the influence of each proposed feature at different combinations while using YOLOv3[12] as the frame-based object detector. HOTA[26] metrics are used to numerically assess the performance of these combinations at the base framerate of 24Hz and an elevated tracking rate of 384Hz with significantly higher temporal resolution. At both tracking rates, results show optimal performance was achieved when combining both proposed features. The two best results under each metric are in bold, for both tracking rates.

| Object Detector | Tracking Rate | Proposed features | | Mask type | | Metrics | | | | | | | |
|---|---|---|---|---|---|---|---|---|---|---|---|---|---|
| | | Bounding Box Refinement | Continuous Event-based Detection | Event-Based | Edge-Based | Mode | HOTA | DetA | AssA | LocA | HOTA(0) | LocA(0) | HOTA-LocA(0) |
| YOLOv3 | 24 Hz | - | - | - | - | *† | 56.6 | 53.0 | 60.8 | 84.2 | 68.1 | 82.0 | 55.9 |
| | | ✓ | | ✓ | | A1 | 59.3 | 56.1 | 62.9 | **87.9** | 68.1 | **86.4** | 58.9 |
| | | | ✓ | ✓ | | A2 | 60.1 | 60.1 | 60.6 | 83.1 | 77.1 | 78.8 | 60.8 |
| | | ✓ | ✓ | ✓ | | **A3** | **64.1** | **64.5** | 63.8 | 86.4 | **77.4** | 83.5 | **64.6** |
| | | ✓ | | | ✓ | B1 | 59.3 | 56.1 | 62.9 | **87.9** | 68.1 | **86.4** | 58.9 |
| | | | ✓ | | ✓ | B2 | 59.9 | 58.2 | 62.1 | 83.4 | 75.4 | 79.6 | 60.0 |
| | | ✓ | ✓ | | ✓ | **B3** | **64.9** | **64.1** | **65.9** | 86.8 | **77.6** | 84.3 | **65.4** |
| | 384 Hz | | | ✓ | | 2† | 52.5 | 50.1 | 55.3 | 84.1 | 63.8 | 81.3 | 51.9 |
| | | ✓ | | ✓ | | A1 | 54.6 | 52.3 | 57.1 | **86.9** | 63.9 | **84.5** | 54.0 |
| | | | ✓ | ✓ | | A2 | 59.3 | 56.0 | 63.1 | 83.1 | **75.8** | 78.6 | 59.6 |
| | | ✓ | ✓ | ✓ | | **A3** | **63.1** | **59.7** | **66.8** | 86.1 | **76.0** | 83.1 | **63.2** |
| | | | | | ✓ | 3† | 51.3 | 48.9 | 54.1 | 84.2 | 62.3 | 81.5 | 50.8 |
| | | ✓ | | | ✓ | B1 | 53.1 | 50.6 | 55.8 | **86.6** | 62.3 | **84.2** | 52.5 |
| | | | ✓ | | ✓ | B2 | 54.6 | 51.3 | 58.4 | 83.4 | 68.8 | 79.2 | 54.5 |
| | | ✓ | ✓ | | ✓ | **B3** | **60.2** | **57.3** | **63.5** | 85.9 | **72.4** | 83.1 | **60.2** |

* Single-modal object detection and tracking using only images (no event data used). † Results from prior work[20].

**Table 2** Results of the ablation study on the influence of each proposed feature at different combinations while using SSD[13] as the frame-based object detector. HOTA[26] metrics are used to numerically assess the performance of these combinations at the base framerate of 24Hz and an elevated tracking rate of 384Hz with significantly higher temporal

resolution. At both tracking rates, results show optimal performance was achieved when combining both proposed features. The two best results under each metric are in bold, for both tracking rates.

| Object Detector | Tracking Rate | Proposed features | | Mask type | | Mode | Metrics | | | | | | |
|---|---|---|---|---|---|---|---|---|---|---|---|---|---|
| | | Bounding Box Refinement | Continuous Event-based Detection | Event-Based | Edge-Based | | HOTA | DetA | AssA | LocA | HOTA(0) | LocA(0) | HOTA-LocA(0) |
| SSD | 24 Hz | - | - | - | - | *† | 69.0 | 67.4 | 70.9 | **89.1** | 77.2 | 87.9 | 67.9 |
| | | ✓ | | ✓ | | A1 | 67.3 | 66.6 | 68.1 | 88.1 | 76.8 | **86.7** | 66.6 |
| | | | ✓ | ✓ | | A2 | 66.8 | 63.9 | **70.2** | 88.0 | 78.7 | 85.0 | 66.9 |
| | | ✓ | ✓ | ✓ | | **A3** | **69.0** | **68.6** | 69.6 | 87.0 | **82.5** | 84.3 | **69.5** |
| | | ✓ | | | ✓ | B1 | 67.3 | 66.6 | 68.1 | 88.1 | 76.8 | **86.7** | 66.6 |
| | | | ✓ | | ✓ | **B2** | **69.2** | 67.7 | **71.0** | 88.3 | 79.9 | 86.0 | **68.7** |
| | | ✓ | ✓ | | ✓ | B3 | 68.5 | **68.1** | 69.0 | 87.4 | **80.2** | 85.3 | 68.4 |
| | 384 Hz | | | ✓ | | 2† | 65.0 | 62.5 | 67.8 | **88.8** | 73.2 | **87.0** | 63.7 |
| | | ✓ | | ✓ | | A1 | 63.0 | 61.1 | 65.1 | 87.3 | 73.0 | 85.1 | 62.1 |
| | | | ✓ | ✓ | | **A2** | **65.9** | 62.0 | **70.2** | 87.9 | **77.6** | 84.6 | **65.6** |
| | | ✓ | ✓ | ✓ | | **A3** | **66.4** | **63.9** | 69.2 | 86.5 | **79.4** | 83.6 | **66.4** |
| | | | | | ✓ | 3† | 62.5 | 60.4 | 64.7 | **88.7** | 70.6 | **86.9** | 61.3 |
| | | ✓ | | | ✓ | B1 | 60.4 | 58.9 | 62.1 | 86.9 | 70.3 | 84.7 | 59.6 |
| | | | ✓ | | ✓ | B2 | 63.8 | 60.2 | 67.7 | 88.0 | 74.1 | 85.4 | 63.2 |
| | | ✓ | ✓ | | ✓ | B3 | 63.3 | 60.7 | 66.1 | 86.3 | 75.3 | 83.6 | 63.0 |

* Single-modal object detection and tracking using only images (no event data used). † Results from prior work[20].

Ref. 26. Further, we note that the output was recorded and saved in the MOTChallenge format[45] where it is used to calculate the final detection and tracking metrics using TrackEval[50], developed by J. Luiten. For compactness, we report the results of only two tracking rates, 24 and 384Hz, given that they should provide both ends of the performance spectrum, where intermediate tracking rates are expected to perform within that range.

*3.4.2 Ablation study results*

The results of the ablation study are presented in Table 1 and Table 2 for the frame-based object detectors YOLOv3[12] and SSD[13], respectively.

In Table 1, results show significant performance advantages when both methods are utilized with either event-mask type. At 24Hz, we notice that the DetA improves from 53% to 64.5% and 64.1% for the event-based (A3) and edge-based (B3) masks, respectively. Similarly, the AssA improves from 60.8% to 63.8% and 65.9% under the event mask types as well, with the final HOTA equal to 64.1% and 64.9%. Further, we observe that the LocA performance of YOLOv3 has significantly improved with the introduction of the BB refinement methods, as indicated by A1 and B1, and the combined methods A3 and B3. Meanwhile, at the higher tracking rate of 384Hz, the combination of the two features presents the most performance gain consistent with the results at 24Hz, showing significant performance gains over our previous work's results (as indicated by modes 2 and 3) under all the metrics. This is highlighted by the substantial improvement in the general HOTA metric values from 52.5% to 63.1% and from 51.3% to 60.2%, when incorporating both methods under the event-based and edge-based masks, respectively. Thus, showing significant performance advantages for optimized object detectors and even more limited performance deterioration at higher temporal resolutions for object tracking. Note that the performance variance at 384 Hz can be attributed to the inconsistent frame-based object detection performance of YOLOv3, which results in frequent detection gaps in the images. As a result, based on our framework's design, these frame-based detections are required to initiate the inter-frame event-based detection process. Therefore, such gaps can be substantially amplified at higher tracking rates. For example, there would be one window frame that contains an image for every 15 window frames that contain only events, at a tracking rate of 384 Hz with a 24 Hz camera framerate. Accordingly, a missed detection in a given image can cause a missed detection, and gaps in the object tracking trajectory, for a total of 16 consecutive window frames. This further emphasizes the importance of the continuous event-based object detection process introduced in this work.

Conversely, our proposed methods had slightly less performance improvement when using SSD as the frame-based object detector, as shown in Table 2. At 24 Hz, we show that the DetA marginally improves from 67.4% to 68.6% and 68.1% for the combined features using an event-based and edge-based mask, respectively, as indicated by modes A3 and B3, whereas the AssA was already high, to begin with, and did not show any meaningful benefit. This is due to the relative simplicity of the dataset used (few occlusions and a low number of objects simultaneously at any time instant), as well as the robust object detection performance of the SSD variant used in this study. Furthermore, the LocA was a bit adversely impacted by either combination of the proposed methods. This demonstrates the limitations of the BB refinement method which can negatively affect the BB precision of a robust object detector such as SSD, as discussed in Sec. 3.2.2. Nevertheless, the more comprehensive metric, HOTA, showed consistent performance in A3, and slightly improved performance using the edge-based mask combined with the continuous event-based detection feature. On the other hand, at 384Hz, the proposed methods showed respectable improvements to the prior results[20]. Both the DetA and the AssA, as well as the general HOTA values, showed improvements, with the A3 configuration yielding the best performance at the tracking rate of 384 Hz. Overall, for SSD, the proposed methods had mixed effects initially but were more beneficial at higher tracking rates.

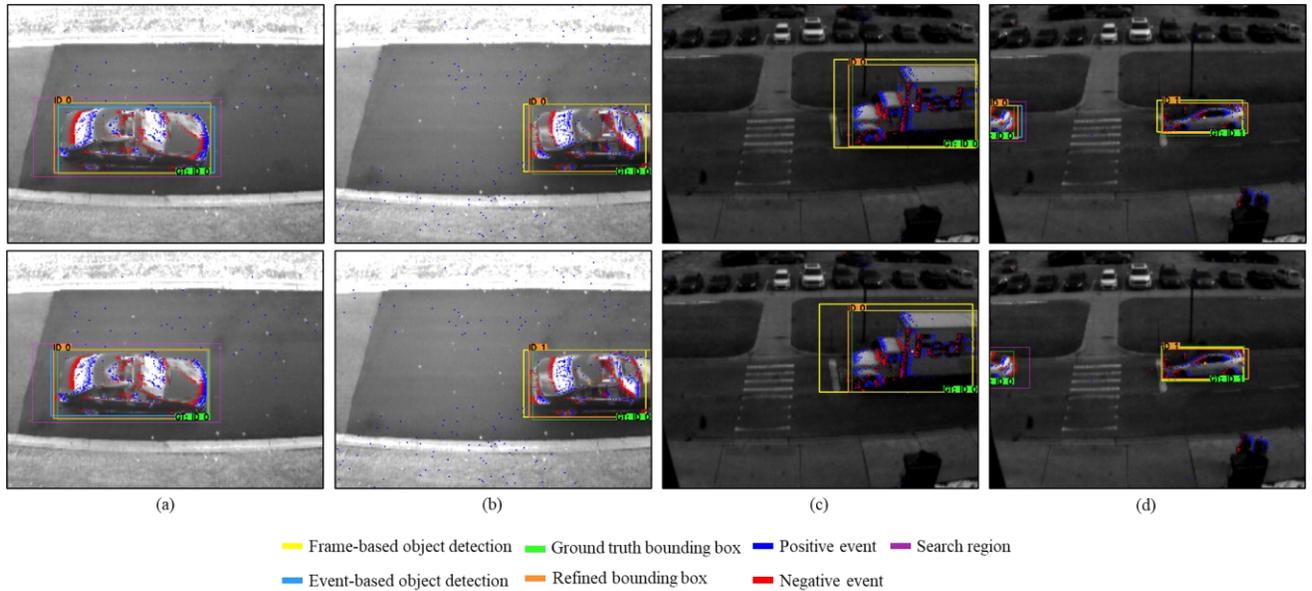

**Fig. 6** Qualitative results of our multi-modal object detection and tracking framework at the baseline tracking rate of 24 Hz at various instances. The top and bottom rows represent the same time instant but for different configurations. (a) and (b) demonstrate the output of YOLOv3 with A3 (top) and B3 (bottom) event-based configurations, whereas (c) and (d) compare the A3 configuration for both YOLOv3 (top) and SSD (bottom). This demonstrates the effects of the proposed methods in combination with our multi-modal framework.

Finally, we provide some qualitative results of our system at the baseline tracking rate of 24 Hz, as shown in Fig. 6, to visually demonstrate the effects of the proposed methods under different configurations and settings. For example, Fig. 6(a) shows an instance where the frame-based object detector used, YOLOv3, missed detecting the vehicle in the scene. Instead, this object, which was previously tracked, is detected using the event-based method for configurations A3 (top) and B3 (bottom), whereas (b) shows the BB refinement process which generates a more accurate BB, relative to the ground truth label, compared to the initial frame-based object detection for the same configuration. In the same manner, Fig. 6(c) and (d) show the output of the A3 tracking configuration for both YOLOv3 (top) and SSD (bottom), showing multiple tracked objects with the application of the bounding box refinement process, in addition to the event-based object detection method that recovered an object in the YOLOv3 configuration (top) but failed in the SSD configuration (bottom) as demonstrated in Fig. 6 (d).

**Table 3** Computational latency analysis of the main stages of the proposed multi-modal object detection and tracking framework on a relative basis, using only a CPU. Event-based methods demonstrate at least an order of magnitude lower latency than the frame-based object detectors' inference times. The estimated total latency refers to the worst-case scenario, where a window frame contains an image of an object that was initially missed, then detected using the event-based method, based on the specified configuration, while refining the resulting bounding box. The event-based methods are estimated per single object with an average bounding box size of 80×45 pixels.

| Stage | Method | Average Latency (ms) |
|---|---|---|
| Frame-based Object Detection | SSD-300[13] (VGG19) | 359 |
| | YOLOv3-320[12] (DarkNet53) | 225 |
| Event Mask Generation | Event-based mask | 0.66 |
| | Edge-based mask | 0.27 |
| Event-based Object Detection | using Event-based mask | 23.7 |
| | using Edge-based mask | 29.0 |
| Additional Methods | Bounding box refinement | 0.37 |
| | Continuous detection* | 23.7-29.0 |
| Total Latency | A3 SSD | 384 |
| | B3 SSD | 389 |
| | A3 YOLOv3 | 250 |
| | B3 YOLOv3 | 256 |

\* Equivalent to the event-based object detection's latency depending on the type of event mask used.

*3.4.3 Computational latency analysis*

To analyze the overall performance and the computational requirements of the system, we provide a simple computational latency analysis for the different core components of the system on a relative basis, as demonstrated in Table 3. Note that all the tests, in our analysis, were implemented and conducted on a CPU, specifically Intel i7-7700HQ. Therefore, a GPU was not used in our testing. Moreover, the event-based methods, at this stage, are not optimized for runtime performance (i.e., no multi-threading or compiled libraries) given that this work was implemented using the scripting language, Python. We also note that, due to their classical design, the computational latency of the presented event-based methods is linearly proportional to the object's size. Accordingly, we present the results per one object for the detection and tracking process, where the average BB size was found to be around 80×45 in these tests. Consequently, this analysis is meant to highlight the difference in the computational requirements and average latencies of the event-based and the frame-based components as well in a relative manner. This leaves significant room for future work in terms of optimizing this framework for better real-time performance, which is not the focus of this work.

Overall, the results in Table 3 show a significant disparity between the latency of the frame-based object detection components and the event-based components. This can be attributed to the scale and complexity of the learned components used. As for the difference in the average latency of both frame-based object detectors, we note that YOLOv3's[12] architecture utilizes the DarkNet-53 feature encoder as its backbone which has 42 million parameters[51]. Meanwhile, SSD-300's architecture uses VGG-19[19] instead, which has over 143 million parameters, thus, justifying the 134 ms variance in their average latency, and the difference in their single-modal object detection and tracking performance (at 24Hz) as demonstrated in Table 1 and Table 2. In contrast, event-based methods presented have shown at least an order of magnitude less latency than their frame-based counterpart. For instance, the initial event mask generation process takes only 0.66 and 0.26 milliseconds, on average, for the event-based and edge-based masks, respectively. This is followed by the inter-frame, event-based, object detection process, described previously in Sec. 3.1, which takes 23.7 ms, on average, when using the event-based mask, and 29 ms when using the edge-based mask. Therefore, the combined, multi-modal approach presented in this paper can enable high-temporal-resolution object detection and tracking results with minimal overhead using a low-framerate camera, compared to the single-modal approach that incorporates a high-framerate camera with DNN-based frame-based object detectors.

As for the computational overhead resulting from the additional methods presented, we notice that the BB refinement process provides a very low-cost solution that delivers noticeable performance improvements with just 0.37 ms of average latency. Meanwhile, the continuous event-based object detection process has an overhead equal to the main inter-frame object detection process, given that they utilize the same procedure. However, this method is only used when a previously tracked object was not detected in a given image, thus resorting to the asynchronous event domain in an attempt to recover it. Accordingly, we estimate the worst-case total latency possible for our multi-modal framework at the baseline camera framerate of 24Hz. This scenario assumes that an object was not initially detected by the frame-based object detector selected, followed by the process of event-based object detection, in addition to the BB refinement process, as represented by the A3 and B3 tracking modes described earlier in this section. The results show that the base performance of YOLOv3, in addition to the event-based methods presented, with minimal computational overhead based on our framework and a total latency of around 250 ms, can compete with the performance of the single-modal, frame-based, tracking-by-detection approach that uses SSD with 359 ms latency (> 100 ms difference).

In summary, the results presented validate our assumptions, especially under more efficient and optimized DNN object detectors such as YOLOv3. Our results showed significant overall object detection and tracking improvements using classical computer vision techniques that leverage event data. The best performance was achieved when combining both proposed methods under either event mask type, especially when using the event-based mask. The outperformance of the event-based mask further proves the benefit of incorporating the polarity data of events in object detection and tracking applications.

**4 Experiment Design**

To further prove the validity of our event-based approaches, we set up and establish an additional experiment to yield real-world distance measurements of vehicle tracking. The experiment is designed to evaluate the vehicle's detection and tracking performance using a high-temporal-resolution positional tracking device. This section describes the design of this experiment, including how the data was collected and processed, as well as the metrics used in our evaluation.

Accordingly, we set up a single-object detection and tracking experiment that incorporates a high-speed LiDAR to get high-temporal-resolution ground truth distance measurements of a vehicle being tracked. The industrial high-speed LiDAR Benewake TF03-100 is used, which can measure distances up to 100m while delivering a distance resolution of 1 cm and an update rate of 1000Hz. Further, it has an estimated error of ±10 cm within 10 m distances and about 1% error at 10m and above. This LiDAR was utilized in 13 out of the 63 collected dataset sequences[20], discussed in Sec. 3.4, which was placed at the end of the street and directed at the vehicle of interest, which is driving towards it at varying speeds and acceleration rates. In the meantime, the hybrid event camera, DAVIS 240, was set up on a building with a high elevation pointing downwards across the street, as described earlier in Sec. 3. Based on our measurements, the vehicle enters the camera's field of view, resembled by the captured scene, at around the 40-meter mark and exits at around 28 meters from the LiDAR.

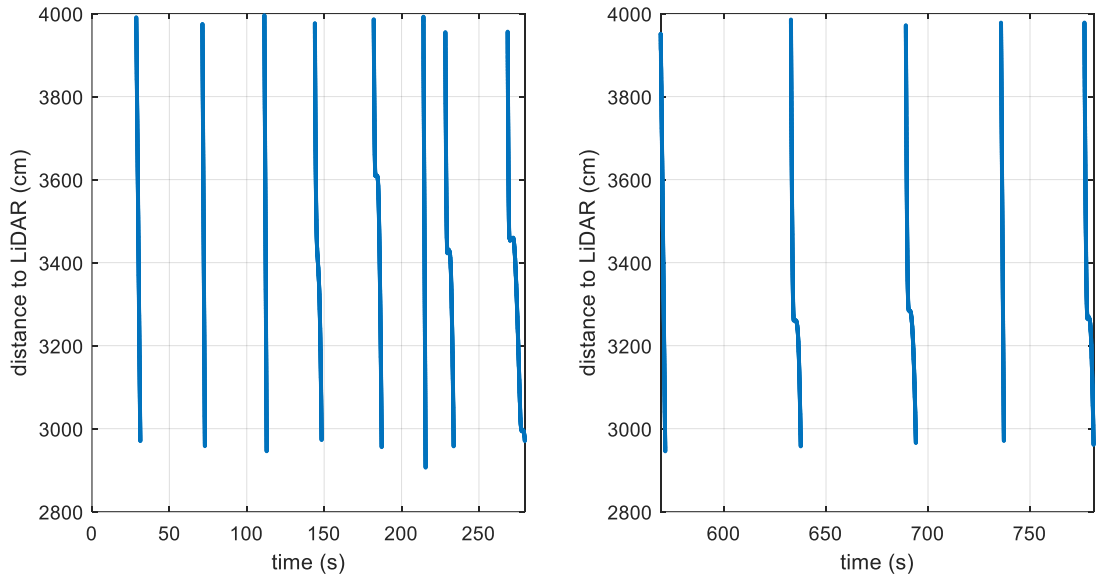

**Fig. 7** Filtered LiDAR ground truth distance data collected with a sampling rate of 1000Hz showing 13 trajectories for the single-object tracking validation experiment. The time between 300 and 500 seconds is removed due to no available tracking data.

### 4.1 Data preprocessing

To make use of the data collected to estimate the performance of our tracking methods, the data from the different modalities must be preprocessed and synchronized to yield useful data.

#### 4.1.1 Ground truth LiDAR data preprocessing

After recording the distance measurements using LiDAR, the gaps due to the sporadic missing data points of the undetected vehicle are initially filled using linear interpolation. Afterward, the data points are resampled to convert from a non-uniform to a uniform sampling rate of 1000Hz, equivalent to a 1 ms difference between every two consecutive data points. Finally, the resulting data is smoothened and filtered using a Chebyshev Type II-type low-pass filter with a passband and a stopband frequency of 10 and 12 Hz, respectively, as well as a passband ripple of 1dB and a stopband attenuation of 80dB. Thus, generating pre-processed ground truth distance measurements with a sampling rate of 1000Hz. The resulting synchronized and filtered LiDAR-based ground truth data is demonstrated in Fig. 7.

*4.1.2 2D Object tracking to distance measurements*

To convert the 2D BBs of the vehicle detections into estimated distance measurements, we use the center-right coordinate $V_{(x,y)}$ of the detection's BB at a given moment, resembling the front of the vehicle. This coordinate can then be used to estimate the distance from the LiDAR. However, due to the different types of lens distortion, namely radial and tangential, such conversion is non-linear. Hence, the removal of lens distortion is critical for accurate positional estimation. This is achieved by the camera calibration process which generates the camera's geometry (also known as intrinsic parameters) as well as the lens distortion models which are used to correct the captured images, as the event coordinates, as shown in Fig. 8(a).

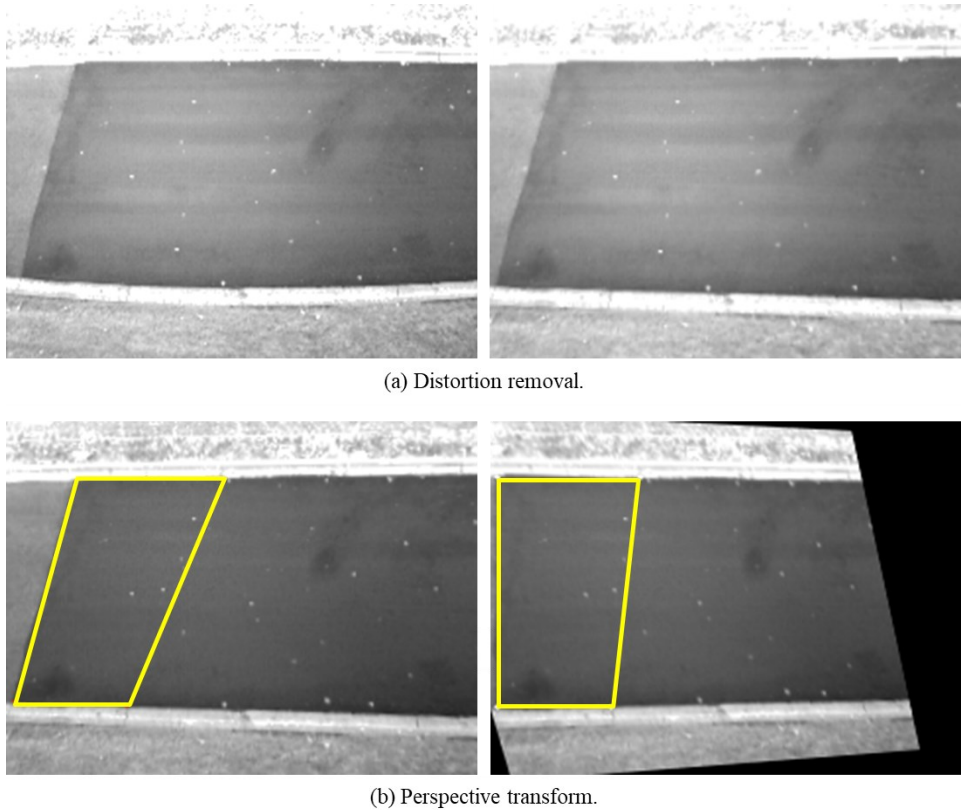

(a) Distortion removal.

(b) Perspective transform.

**Fig. 8** Demonstration of the camera calibration process and bird's eye perspective transform which are required for accurate object tracking and positional measurements. (a) the distortion is removed from the captured image using the DAVIS 240c sensor, (b) then the scene's perspective is transformed to provide a bird's eye view of the region of interest.

Afterward, to enable a linear pixel position-to-distance conversion, the bird's eye-view perspective is a common approach to object position estimation, especially in traffic surveillance applications. However, due to the geometry of the road relative to the camera and its elevation level, a perspective transform is necessary to yield a linear conversion. Accordingly, the perspective transform matrix $M$ of size $3 \times 3$ is generated using 4 initial points (pixels) on the image, along with the desired final coordinates for each of them, to produce a bird's eye view perspective, as shown in Fig. 8(b). Moreover, the 4 chosen points are of known real-world distance measurements, which enable the final conversion from a pixel coordinate, on the x-axis, to a distance-per-pixel value.

We note that this process is performed initially to yield the required transformation matrices where the tracking point $V_{(x,y)}$, which refers to the center-right coordinate of the detected vehicle, is initially undistorted, then transformed using the equation

$$Z_{(x,y)} = \left( \frac{M_{(0,0)}V_x + M_{(0,1)}V_y + M_{(0,2)}}{M_{(2,0)}V_x + M_{(2,1)}V_y + M_{(2,2)}}, \frac{M_{(1,0)}V_x + M_{(1,1)}V_y + M_{(1,2)}}{M_{(2,0)}V_x + M_{(2,1)}V_y + M_{(2,2)}} \right), \tag{1}$$

where $V_x$ and $V_y$ and the undistorted x and y coordinates of the tracking point $V_{(x,y)}$, $M$ is the perspective transform matrix generated earlier, and $Z_{(x,y)}$ is the final transformed pixel position. Note that the center-right coordinate, $V_{(x,y)}$, actually resembles the front of the vehicle. Accordingly, the x-coordinate of the resulting point $Z_{(x,y)}$ is converted to the estimated distance to the LiDAR by normalizing it (subtracting from and dividing by the frame's width in pixel count) and then multiplying by the measured distance-per-pixel (~4.4 cm/pixel). Finally, the resulting point is offset by the estimated distance between the frame's side to the LiDAR (28m). This process is applied to all the tracking points produced by our methods presented in Sec. 3.

*4.1.3 Temporal synchronization*

The system clocks of the different computers are typically synchronized using an online global clock, however, only to the order of seconds at best. Moreover, in our data collection procedure, a convenient network-based time synchronization method was not feasible due to the significant distance between both the camera and the LiDAR, eliminating the possibility of conveniently connecting the data collection computers to enable temporal synchronization. This presents a challenge for our high-temporal-resolution tracking data when evaluating it according to the ground truth distance. Therefore, we intuitively achieve temporal synchronization by using the previously labeled 2D ground truth BBs, generated at a high rate of 384Hz[20], for the 13 sequences involved, and manually synchronizing them with a resampled version of the filtered LiDAR data collected at 1000Hz presented in Sec. 4.1.1, thus minimizing the temporal difference as much as possible. Accordingly, we have found the temporal synchronization difference to be 0.719 seconds between the system clocks of both devices. This value is subsequently used to offset all of the generated vehicle tracking data and enable a millisecond level of synchronization.

*4.2 Experiment Parameters and Metrics*

*4.2.1 Vehicle detection and tracking configurations*

Using the optimal configuration, found in our ablation study in Sec. 3, which combines both improvements proposed, we evaluate the distance estimation results at different uniform rates of 24, 50, 100, 200, and 500Hz, using the tracking modes A3 and B3 described earlier, in combination with the frame-based object detectors YOLOv3[12] and SSD[13]. We match the ground truth LiDAR data to the different tracking rates by resampling it at the defined uniform rates, while only keeping the tracking data where the vehicle is in the camera's field of view till just before it starts leaving the scene (i.e., the front of the vehicle no longer visible in the frame). Then, the resulting outliers are manually removed, if any, which are due to the interpolation required in the resampling process of ground truth data, to yield an identical number of possible tracking points at each tracking rate for proper and accurate evaluation.

*4.2.2 Evaluation metrics*

The results of this experiment are evaluated using several error metrics, including the median absolute error, the median relative tracking error, and the root-mean-square error (RMSE). The metrics are estimated using only the successfully detected points, which are highlighted by the successful detection rates calculated for each tracking configuration. The median error metrics are chosen due to their robustness to outliers in comparison to the mean error metrics. Such outliers can result due to the resampling and temporal synchronization errors, which affect the results but without validity. Further, we note that this experiment does not account for any additional object

Table 4 Vehicle detection and tracking validation experiment results under various detection configurations and tracking rates. A summary of different error metrics is presented, along with the successful detection rates and temporal synchronization errors. We demonstrate that our presented methods successfully leverage event data to enhance vehicle detection and tracking performance at various tracking rates.

| Frame-based Object Detector | Tracking Rate (Hz) | Event-based Detection Mode | Metrics | | | | |
|---|---|---|---|---|---|---|---|
| | | | Median Abs. Error (cm) | Median Relative Error (%) | RMSE (cm) | Successful Detection Rate (%) | Mean Temporal Synchronization Error (s) |
| YOLOv3 | 24 | N/A* | 21.2 | 0.66% | 31.0 | 66.3% | |
| | | A3 | 11.1 | 0.33% | 27.8 | **81.2%** | 0.0112 |
| | | B3 | 11.4 | 0.34% | 22.9 | 79.6% | |
| | 50 | A3 | 9.2 | 0.28% | 17.9 | 80.8% | 0.0043 |
| | | B3 | 10.4 | 0.31% | 19.0 | 76.2% | |
| | 100 | A3 | 7.2 | 0.22% | 15.4 | 81.1% | 0.0016 |
| | | B3 | 8.9 | 0.27% | 17.1 | 75.4% | |
| | 200 | A3 | 6.7 | 0.20% | 14.3 | 81.0% | 0.0013 |
| | | B3 | 8.4 | 0.25% | 16.3 | 75.2% | |
| | 500 | A3 | **6.4** | **0.19%** | **13.6** | 81.1% | **0.0004** |
| | | B3 | 8.1 | 0.25% | 15.9 | 74.9% | |
| SSD | 24 | N/A* | 6.6 | 0.20% | 59.9 | 91.2% | |
| | | A3 | 7.2 | 0.22% | 21.7 | **91.3%** | 0.0112 |
| | | B3 | 7.2 | 0.22% | 21.7 | **91.3%** | |
| | 50 | A3 | 6.2 | 0.18% | 14.8 | 90.5% | 0.0043 |
| | | B3 | 6.4 | 0.20% | 14.8 | 85.9% | |
| | 100 | A3 | 5.2 | 0.16% | 11.4 | 90.3% | 0.0016 |
| | | B3 | 5.5 | 0.17% | 12.7 | 84.3% | |
| | 200 | A3 | 5.0 | **0.15%** | 11.2 | 90.5% | 0.0013 |
| | | B3 | 5.2 | 0.16% | 11.9 | 83.7% | |
| | 500 | A3 | **4.8** | **0.15%** | **10.0** | 90.4% | **0.0004** |
| | | B3 | 5.0 | **0.15%** | 11.5 | 83.3% | |

* Object detection and tracking using image frames only (no event data used). Best values are highlighted in **bold**.

detections besides the vehicle being tracked. These detections are ignored. However, missed detections (false negatives) would affect the vehicle detection success rates. Finally, we provide the mean of the temporal synchronization errors at each rate to provide insight into their effects on tracking performance, if such correlation occurs.

## 5 Results and Discussion

The results of the validation experiment are presented in Table 4. To begin with, we can notice that our presented event-based methods, in combination with our general framework, are advantageous to the vehicle detection and tracking process as highlighted in the lowest tracking rate. In terms of successful detection rates, at 24Hz (equivalent to the image frames sampling rate), the vehicle detection rate improved significantly from 66.4% to 81.3% when using our event-based object detection recovery presented earlier, compared to only frame-based detection when using YOLOv3, further validating our prior assumptions. Similarly, detection results improved, albeit marginally, by 0.1% when using SSD as the frame-based object detector. This is due to the selected SSD variant being a more accurate and stable object detector. Meanwhile, at higher tracking rates, the event-based mask approach A3 has increasingly outperformed the edge-based mask approach B3, ending with a difference of 6.2% for YOLOv3 and 7.1% for SSD at 500 Hz., compared to a negligible difference at 24 Hz.

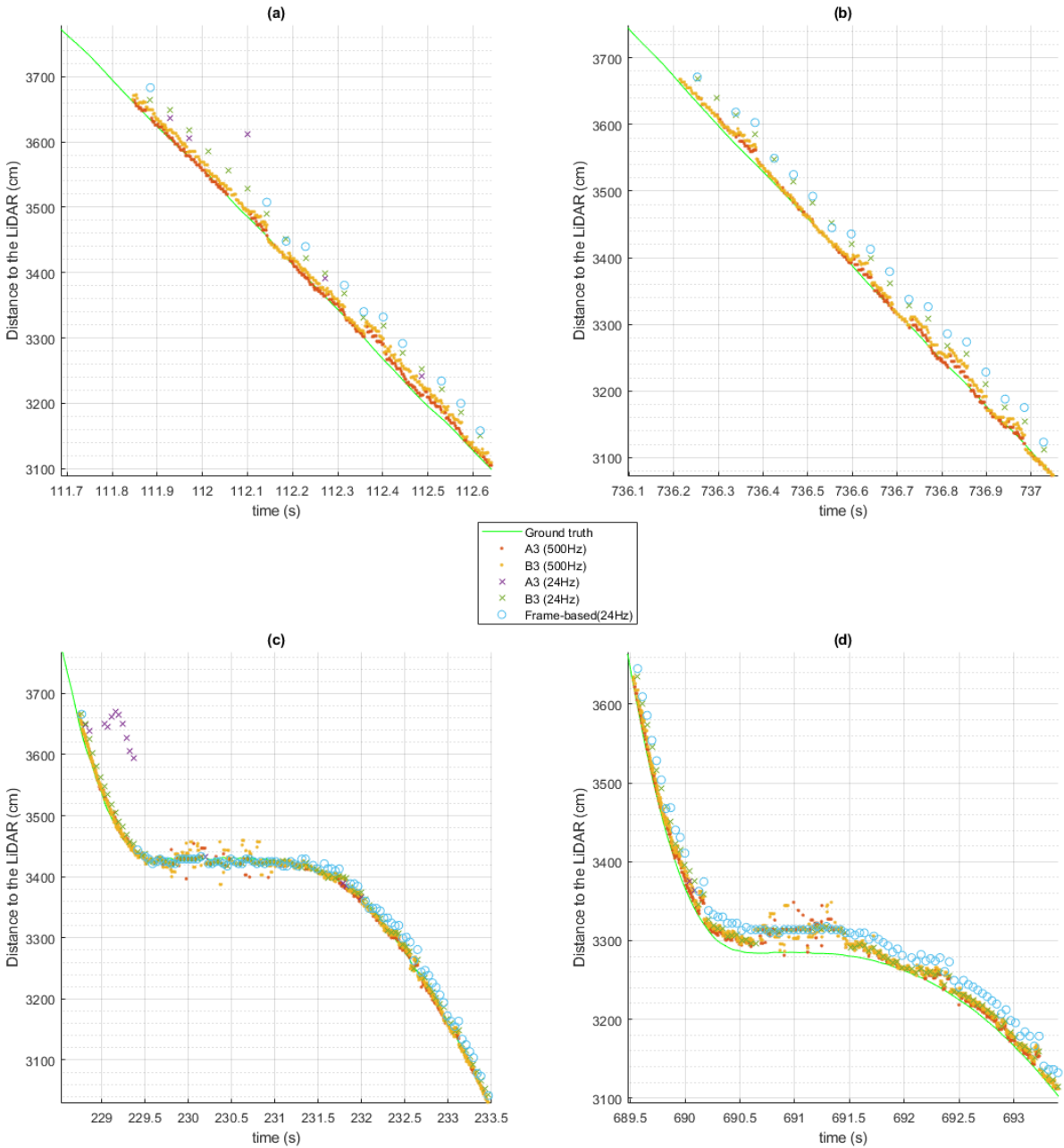

**Fig. 9** Demonstration of the distance estimation tracking results of 4 selected trajectories at the tracking rates 24Hz, and 500Hz, using tracking modes A3 and B3, compared to the ground truth LiDAR data, as well as the frame-based only approach. All results are presented using YOLOv3 which underperformed the other frame-based object detector but benefitted the most from our proposed event-based methods.

As for the main error metrics, we observe that modes A3 and B3 have presented significant benefits when combined with YOLOv3 at 24 Hz, but with mixed results for SSD at the same rate. Nevertheless, at higher tracking rates, the error magnitudes improve substantially for either object detector, where the best multi-modal tracking mode, A3, at 500 Hz resulted in a median absolute error of 6.4 cm and 4.8 cm, for YOLOv3 and SSD, respectively.

Interestingly, we observe that the error rates get consistently lower as we increase the tracking rates, with a negligible deterioration in object detection performance, when using either frame-based object detectors with any event-based object detection mode. This can be partially attributed to the BB refinement process which improves the accuracy of the point $V_{(x,y)}$ that represents the front of the vehicle, where if given more tracking points, it would consistently improve on each. Another reason could be the higher average temporal synchronization error, indicated at the lower tracking rates, which is primarily due to the resampling and synchronization errors of the ground truth data collected at 1000Hz when downsampled to lower tracking rates.

Moreover, in Fig. 9, we plot the results for 4 selected trajectories (sequences 3, 12, 7, and 11) out of the 13 presented in Fig. 7. The results of the modes A3 and B3, at the tracking rates of 24Hz and 500Hz, are compared to the ground truth data and the frame-based-only tracking results at 24Hz, using YOLOv3 as the frame-based object detector. The selected trajectories represent different vehicle acceleration rates, where the vehicle moves with a consistent speed in trajectories (a) and (b), whereas it comes to a full stop before accelerating again in trajectories (c) and (d). Similarly, the results demonstrate the feasibility of high-temporal-resolution tracking, as well as the improvement to the frame-based method at 24Hz. Nonetheless, we notice some variance in the trajectory of the vehicle in (c) and (d) for both proposed methods at 500 Hz, in contrast with the tracking results at 24 Hz. This variance is caused by the low number of events generated when the vehicle is still when reaching a stop. The lack of motion causes a lower number of events to be generated. Therefore, the event-based detection and tracking process is affected by the events generated due to noise or the lack of events generated altogether, causing the estimated distance to be unstable, and negatively affecting the correct detection rates. This further highlights the limitations of event cameras in static scenes and stresses the importance of filtering techniques, which are not considered in this study, to improve the event data's signal-to-noise ratio. Likewise, our framework is also affected by a degraded image frame input in conditions such as continuous low-light or motion blur, which the inter-frame event-based detection process is dependent on, at least initially. This case, however, is not considered in this work due to the lack of proper data for evaluation and testing but can be addressed in future works.

In summary, consistent with the results presented in Sec 3.4, the tracking configuration mode A3, which entails an event-based mask along with both improvement methods presented, consistently provides the best results under the varying tracking rates and either frame-based object detectors. Overall, we conclude that this experiment is successful. Results show tracking results within the ground truth data's margin of error, at various temporal resolutions and tracking rates. Thus, reaffirming the capabilities and advantages of incorporating event data with proper event-based techniques to improve the performance of a frame-based framework by applying low-cost, classical image processing and computer vision techniques on the asynchronous, and high-temporal-resolution event data.

# 6 Conclusion

In this work, we have presented an improved, high-temporal-resolution object detection and tracking framework using a combination of frame and event-based methods. Building on our prior work[20], we have introduced two event-based methods that further enhance the robustness and accuracy of the detection and tracking framework. These methods are event-based BB refinement and continuous event-based object detection and recovery. Using a labeled MOT vehicle dataset with HOTA metrics, an ablation study was conducted, which showed that the two methods combined provide significant performance gains outperforming frame-based and prior approaches alike, at tracking rates from 24 to 384Hz. The results show that these event-based methods, in combination with optimized and real-time object detector models such as YOLOv3, can benefit substantially by incorporating the asynchronous and high-temporal-resolution event data in a multi-modal approach. More specifically, these methods can reduce the effects of intermittent frame-based object detection with various infrequent missed detections, and improve the precision of a detection's bounding box, with minimal computational overhead. This was demonstrated by the absolute improvements of 11.5% in the DetA metric, and 7.5% in the overall HOTA metric, compared to the single-modal frame-based approach at 24 Hz using YOLOv3. Nevertheless, these approaches are still susceptible to false detections (i.e., false positives) produced by the frame-based object detectors, which are given higher confidence due to the presumed robustness of these models. This is an indirect result of classical methods that are not quite dynamic and require a decent amount of handcrafting. Instead, some learned event-based methods can be explored to replace some of the proposed components of our presented framework with a more dynamic approach toward handling noise or signal degradations resulting from either modality.

Furthermore, a validation experiment was designed and conducted to demonstrate the usefulness of our hybrid framework using real-world values. A high-speed LiDAR was used to collect ground truth distance measurements for vehicle tracking at a 1000Hz sampling rate. The vehicle detection and tracking results were generated at various temporal rates, including 24Hz (equal to the framerate of the APS) as well as 50, 100, 200, and 500Hz high-resolution tracking rates. The tracking results were assessed using different error metrics and overall detection success rates. Results showed that high-temporal-resolution tracking is feasible with output within the ground truth data's margin of error, as well as very high successful detection rates, yielding a true high-resolution tracking output by utilizing event data appropriately.

Overall, this work demonstrates the effectiveness and capabilities of event-based vision, and how well it can complement frame-based vision for different computer vision tasks. The properties of this sensing modality provide great potential that requires proper methods to fully utilize it. Future work potential includes replacing some of the presented classical and hand-crafted event-based components with learned ones to achieve a more dynamic and robust performance under various challenging scenarios for both modalities, such as non-static scenes with ego-motion for the event-based methods; and low-light or motion blur for the frame-based methods. Nevertheless, that would require larger amounts of labeled event and multi-modal datasets, and possibly other event representations. Furthermore, this work can be tailored to on-vehicle cameras that are essential in autonomous vehicles and automated driving, with further development and applicable datasets. These approaches would require the ability to differentiate between foreground and background events to enable robust object detection and tracking performance.

*Disclosures*

The authors declare no conflict of interest.

*Acknowledgments*

The authors of this paper would like to thank Mariana A. Al Bader, for her great support in annotating and labeling the collected dataset, and Ella Reimann, for her help and assistance in the data collection experiment.

*Code, Data, and Materials Availability*

The labeled dataset used in this work is available at http://sar-lab.net/event-based-vehicle-detection-and-tracking-dataset/.

**Zaid El Shair** is a Ph.D. candidate at the University of Michigan-Dearborn. He received his B.Sc. degree in computer engineering at the Princess Sumaya University for Technology, Amman, Jordan, in 2018, and his M.S.E. degree in computer engineering as well from the University of Michigan-Dearborn, MI, in 2019. He is the author of several papers in different domains such as computer vision and embedded systems. His current research interests include event-based vision techniques for robust object detection and tracking and other computer vision-related applications.

**Samir A. Rawashdeh** received his B.Sc. degree in electrical engineering from the University of Jordan, Amman, Jordan in 2007, and his M.S.E.E. and Ph.D. degrees in electrical engineering from the University of Kentucky, Lexington, KY, in 2009 and 2013, respectively. He is currently an Associate Professor in the Electrical and Computer Engineering department at the University of Michigan – Dearborn, which he joined in 2014. His research activities and interests include robot perception, embedded systems, and smart health.